\documentclass{article}
\PassOptionsToPackage{numbers, compress}{natbib}

\usepackage{neurips_2024}

\usepackage[utf8]{inputenc} 
\usepackage[T1]{fontenc}    
\usepackage{url}            
\usepackage{booktabs}       
\usepackage{amsfonts}       
\usepackage{nicefrac}       
\usepackage{microtype}      
\usepackage{xcolor}         
\usepackage{float}
\usepackage{graphicx}
\usepackage{hyperref}       
\usepackage{caption}
\usepackage{lipsum}  
\usepackage{float}

\title{Are EEG-to-Text Models Working?}

\begin{document}
\maketitle

\begin{abstract}
This work critically analyzes existing models for open-vocabulary EEG-to-Text translation. We identify a crucial limitation: previous studies often employed implicit teacher-forcing during evaluation, artificially inflating performance metrics. Additionally, they lacked a critical benchmark – comparing model performance on pure noise inputs. We propose a methodology to differentiate between models that truly learn from EEG signals and those that simply memorize training data. Our analysis reveals that model performance on noise data can be comparable to that on EEG data. These findings highlight the need for stricter evaluation practices in EEG-to-Text research, emphasizing transparent reporting and rigorous benchmarking with noise inputs. This approach will lead to more reliable assessments of model capabilities and pave the way for robust EEG-to-Text communication systems. Code is available at \href{https://github.com/NeuSpeech/EEG-To-Text}{https://github.com/NeuSpeech/EEG-To-Text}

\end{abstract}

\section{Introduction}

The field of brain-computer interfaces (BCIs) has witnessed significant progress, particularly in translating electroencephalography (EEG) signals into written text. This technology offers immense potential for individuals with communication disabilities by enabling direct conversion of brain activity into language. The rapid progress in this area has stimulated extensive research aimed at refining the accuracy, speed, and overall usability of EEG-to-Text models, with the ultimate aim of facilitating smooth and effective communication for a wider population.\let\thefootnote\relax
\footnote{*Equal contribution †Corresponding author \\
Corresponding author email: \texttt{xionghui@hkust-gz.edu.cn}, \texttt{whlee@khu.ac.kr} \\
}

However, despite advancements, existing studies on EEG-to-Text translation exhibit methodological shortcomings that limit the validity and reliability of their findings. One major concern is the use of implicit teacher-forcing during model evaluation~\cite{wang2022open_aaai_eeg2text}. This technique can artificially inflate performance metrics, resulting in an inaccurate assessment of a model's true capability. Additionally, previous studies ~\cite{wang2022open_aaai_eeg2text,duan2023dewave_brain2text,xi2023unicorn} have not compared EEG inputs with pure noise, a crucial benchmark for evaluating the effectiveness of EEG-based communication systems. While prior research has reported high-performance metrics using EEG data (e.g., BLEU scores~\cite{papineni2002bleu_orig}), they have not compared these results to models fed with pure noise of the same length. This omission is critical because similar performance on noise suggests the model is not effectively learning from EEG inputs and might be memorizing text instead.

Addressing these limitations, Yang \textit{et al.}~\cite{yang2024decode} compared their model with noise and magnetoencephalography (MEG) inputs, but their analysis did not encompass mainstream EEG-to-Text models ~\cite{wang2022open_aaai_eeg2text}. This paper addresses this gap by conducting a comprehensive examination of existing EEG-to-Text models, particularly focusing on their performance in open-vocabulary settings. Our analysis reveals a significant deficiency in previous evaluation methodologies, specifically the lack of comparison between EEG inputs and pure noise. This comparison is essential to accurately assess a model's ability to extract meaningful information from EEG signals, as it differentiates genuine learning from simple memorization.

Through our comparative analysis, we demonstrate that current EEG-to-Text models achieve similar or even better performance on pure noise compared to actual EEG inputs. This finding raises concerns about the true learning capabilities of these models and underscores the need for more transparent and rigorous evaluation methodologies within the field.

The main contributions of this paper are summarized as follows:
\begin{enumerate}
    \item We introduce a novel methodology to assess whether EEG-to-Text models are genuinely learning from EEG inputs or simply memorizing patterns.
    \item By shedding light on the learning behavior of existing models, our findings offer valuable insights for future research and development efforts in EEG-based communication systems.
    \item We emphasize the importance of improved transparency and rigorous benchmarking practices in EEG-to-Text research, paving the way for more reliable and trustworthy research in this rapidly evolving field.
\end{enumerate}

\begin{figure}[t]
    \centerline{\includegraphics[width=\columnwidth]{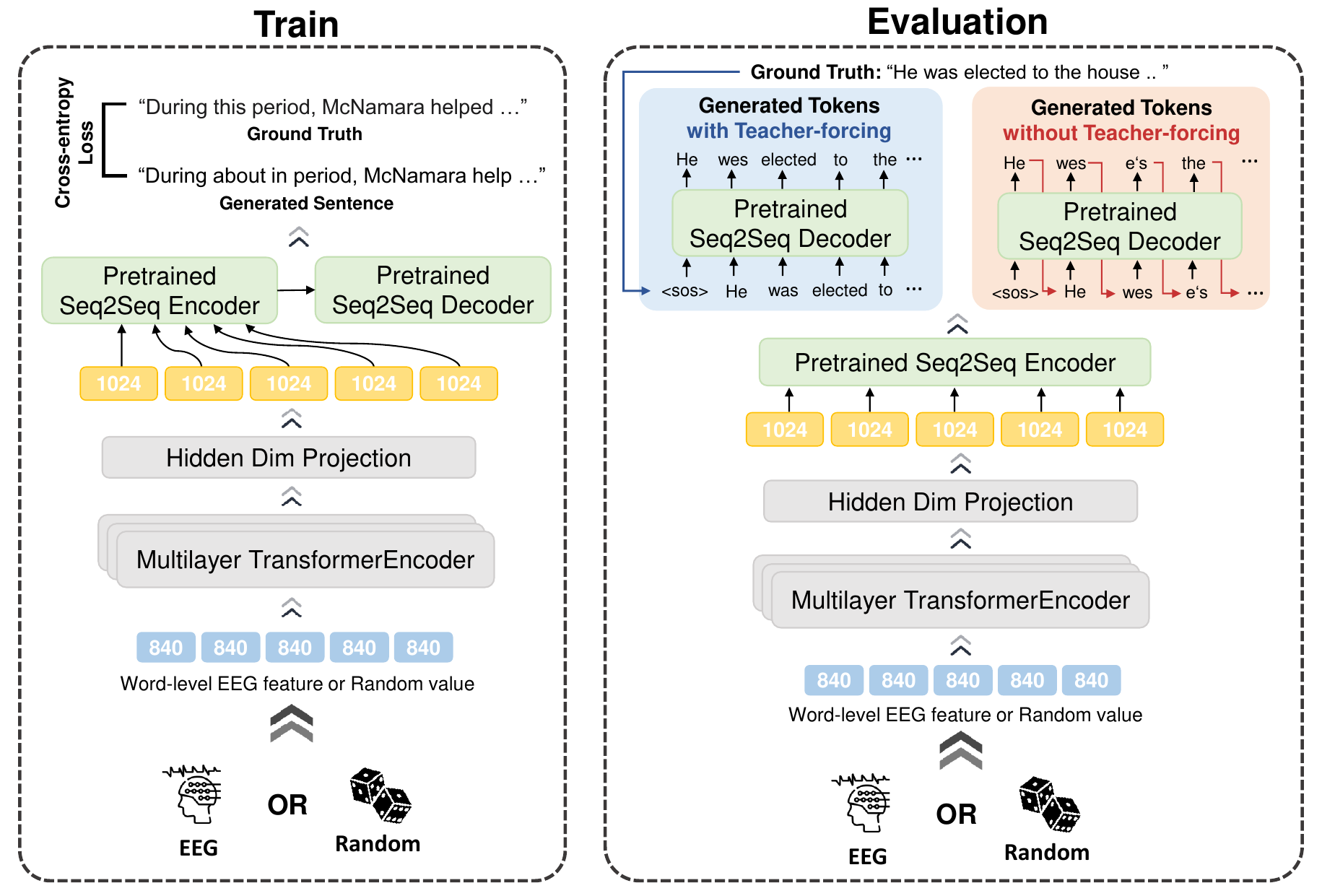}}
    \caption{Schematic illustration of the pipeline for a comprehensive assessment of EEG-to-Text models across four distinct training and evaluation setups \cite{wang2022open_aaai_eeg2text}. These setups explore various combinations of training with either EEG data or random noise as input, followed by evaluation on the same type of data. This approach reveals how models perform under different input conditions. Each setup is further divided to show the influence of teacher-forcing on text generation.}
    \label{fig:figure1}
\end{figure}

\begin{table}[t]
\centering
\renewcommand{\arraystretch}{1.15} 
\caption{Number of training, development, and testing samples for each task within the ZuCo datasets. SR: Normal Reading (movie reviews), NR: Normal Reading (Wikipedia), TSR: Task Specific Reading (Wikipedia).}
\label{sample-count-table}
\begin{tabular}{ccccc} 
\toprule
\textbf{Reading task} & \begin{tabular}[c]{@{}c@{}}\textbf{\#Training }\\\textbf{samples}\end{tabular} & \begin{tabular}[c]{@{}c@{}}\textbf{\#Development }\\\textbf{samples}\end{tabular} & \begin{tabular}[c]{@{}c@{}}\textbf{\#Testing }\\\textbf{samples}\end{tabular} & \begin{tabular}[c]{@{}c@{}}\textbf{\#Total }\\\textbf{samples}\end{tabular}  \\ 
\midrule
SR v1.0               & 3609                                                                           & 467                                                                               & 456                                                                           & 4533                                                                         \\
NR v1.0               & 2645                                                                           & 343                                                                               & 350                                                                           & 3343                                                                         \\
NR v2.0               & 3697                                                                           & 458                                                                               & 392                                                                           & 4547                                                                         \\
TSR v1.0              & 4456                                                                           & 522                                                                               & 601                                                                           & 5579                                                                         \\
\bottomrule
\end{tabular}
\end{table}

\section{Materials and Methods}
\label{headings}

\subsection{Task Definition}

In this replication study, we decoded sentences $\mathcal{S}$ using a sequence of word-level EEG features $\mathcal{E}$ within a sequence-to-sequence (seq2seq) framework. Specifically, we employed EEG-text pairs <$\mathcal{E}, \mathcal{S}$> collected during natural reading tasks, as detailed in Section 2.2.

\subsection{ZuCo Datasets}

We followed the approach of Wang and Ji \cite{wang2022open_aaai_eeg2text} and utilized two publicly available datasets: ZuCo 1.0 \cite{hollenstein2018zuco} and ZuCo 2.0 \cite{hollenstein2020zuco}. These datasets consist of EEG signals and eye-tracking data collected while participants read movie reviews and Wikipedia articles naturally. Essentially, ZuCo provides a collection of EEG recordings and corresponding eye movement data captured during natural reading tasks. The reading materials encompass movie reviews and Wikipedia articles.
From the EEG signals, we extracted 840 features. These features were segmented based on eye fixations for each word encountered during reading. Each feature represents the average amplitude obtained after applying a Hilbert transform across eight predefined frequency bands: theta1 (4-6 Hz), theta2 (6.5-8 Hz), alpha1 (8.5-10 Hz), alpha2 (10.5-13 Hz), beta1 (13.5-18 Hz), beta2 (18.5-30 Hz), gamma1 (30.5-40 Hz), and gamma2 (40-49.5 Hz). Sentences containing NaN values were excluded from the analysis. Subsequently, we partitioned the data for each reading task into three sets: 80$\%$ for training, 10$\%$ for development, and 10$\%$  for testing. Importantly, these partitions were based on unique sentences to ensure no overlap between training, development, and testing data. Table~\ref{sample-count-table} provides a detailed breakdown of the final sample sizes within the dataset. We leveraged the top-performing task combinations identified by Wang and Ji \cite{wang2022open_aaai_eeg2text} for both training and evaluating our models. These combinations included SR v1.0 + NR v1.0 + TSR v1.0 and SR v1.0 + NR v1.0 + NR v2.0 (refer to the original paper by Wang and Ji \cite{wang2022open_aaai_eeg2text} for detailed descriptions of these task combinations).

\subsection{EEG-to-Text Decoding Evaluation}
\paragraph{EEG-to-Text Decoding}
Our analysis builds upon the EEG-to-Text model proposed by Wang and Ji \cite{wang2022open_aaai_eeg2text}. This model is a combination of two key components: 1) An encoder module consists of a multi-layered transformer encoder with six layers and eight attention heads. It processes word-level EEG features as input and generates embeddings through a feed-forward network; 2) A pretrained BART model takes the embeddings from the encoder and utilizes its decoder to generate text sequences by predicting tokens from its vocabulary. The model is trained to minimize text reconstruction cross-entropy loss, essentially aiming to accurately reproduce the original text based on the provided EEG features.

The original approach by Wang and Ji \cite{wang2022open_aaai_eeg2text} employed teacher-forcing during training, raising questions about its impact on generalizability and true model capabilities\footnote{\href{https://github.com/MikeWangWZHL/EEG-to-Text}{https://github.com/MikeWangWZHL/EEG-to-Text}}. To ensure reliable testing and explore the true learning capabilities of the models, we extended our experiments beyond BART. We included two additional pretrained seq2seq models based on the transformer architecture, including PEGASUS \cite{zhang2020pegasus}, and T5 \cite{raffel2020exploring}. All models were used in their LARGE variant only. To assess the model's ability to learn from EEG data and address concerns about teacher-forcing, we designed four distinct training and testing scenarios:

\begin{enumerate}
    \item EEG (training and testing): In this scenario, the model is trained and tested entirely on word-level EEG features. This setup directly assesses the model’s capability to genuinely generate text based on EEG data. It reveals whether the model can extract meaningful information from the EEG signals and translate it into coherent language.
    \item Random (training and testing): The model is trained and tested using random noise instead of actual EEG features. This establishes a crucial baseline performance level. By comparing the model's performance on random noise to its performance on EEG data (in scenario 1), we can gain insights into whether the model is learning from the EEG information or simply memorizing patterns. If the model performs equally well on both EEG and random noise, it suggests it might not be effectively utilizing the EEG data.
    \item EEG (training) + Random (testing): The model is initially trained using EEG features. However, during evaluation, it is tested on random noise data. This setup helps us understand how well the model generalizes its learned patterns from EEG data to unseen data, even if that data is not actual EEG.
    \item Random (training) + EEG (testing): The model is trained on random noise, essentially providing it with no meaningful information. However, during evaluation, it is tested on actual EEG features. This scenario serves as a control to assess the model's ability to generate any meaningful text output in the complete absence of relevant training data. 
\end{enumerate}

By analyzing the model's performance across these four scenarios, we can gain valuable insights into how effectively the models learn from EEG data and address the concerns raised about teacher-forcing techniques. This comprehensive evaluation approach allows us to differentiate between genuine learning and potential memorization of training patterns.

\paragraph{Training}
All models were trained on Nvidia RTX 4090 GPUs using a batch size 32 and a learning rate of $2e^{-5}$. Training was conducted for 30 epochs for each model. The model with the lowest loss on the development dataset was chosen as the best-performing model. Stochastic gradient descent was employed as the optimizer for all training processes.

\paragraph{Evaluation}
This study investigated the influence of teacher-forcing on text generation from EEG features during the evaluation phase. teacher-forcing is a training technique commonly used in recurrent neural networks (RNNs) \cite{cho2014learning}. It involves using the ground truth (correct output) from the training data as the input for the next time step, rather than relying solely on the model's own predictions. This approach can accelerate learning and improve training performance because it ensures the model receives accurate information during the training process.
However, in the inference phase (when the model generates new text), teacher-forcing is not typically used.  Instead, the model generates predictions based on its own outputs from previous steps. This might not reflect the model’s capabilities in real-world scenarios. To assess the impact of teacher-forcing, we evaluated the model’s performance under two conditions: 1) With teacher-forcing: here, the model inferences with the actual output from the previous time step during evaluation, replicating the approach used in the original study \cite{wang2022open_aaai_eeg2text}; 2) Without teacher-forcing: This condition tested the model's ability to generate text independently. The model relied solely on its predictions from previous steps, without any input of the actual ground truth. 
To mitigate repetition in the generated text without teacher-forcing, we employed beam search with a beam size of 5, along with a repetition penalty of 5 and a restriction on the size of repeated n-grams (sequences of words) to 2. Following prior research \cite{wang2022open_aaai_eeg2text, duan2023dewave_brain2text,feng2023aligning, zhou2023belt}, we evaluated the model’s translation performance on the ZuCo dataset using established natural language processing (NLP) metrics, including BLEU \cite{papineni2002bleu_orig}, ROUGE \cite{lin2004rouge_orig}, and WER \cite{klakow2002testing}. These metrics provide quantitative measures of how well the generated text aligns with the reference text from the dataset.

\section{Results and Discussion}
\label{others}

Our analysis yielded two significant findings regarding EEG-to-Text models and the impact of teacher-forcing:
The use of teacher-forcing during evaluation resulted in a substantial performance boost (at least a three-fold increase) compared to models evaluated without teacher-forcing (Tables ~\ref{Results_TSRver} and ~\ref{Results_NRver}). This is evident in Table 4, where models consistently generated text closely resembling the ground truth when teacher-forcing was used, regardless of training or evaluation data type. In contrast, models without teacher-forcing frequently began their outputs with "He was..." irrespective of the actual content. This highlights the significant influence of teacher-forcing on reported performance metrics, suggesting previous studies may have overestimated model capabilities due to this technique. 
We observed consistent evaluation results across models trained on either EEG or random input data, regardless of the evaluation data type (EEG or random). This consistency suggests that the pretrained seq2seq models primarily rely on the label tokens provided during training, not the actual content of the input data (EEG or random). Furthermore, the T5 model consistently achieved BLEU-1 scores approximately 3$\%$ higher than other models, suggesting its text generation capabilities are heavily influenced by its pretrained architecture. 
The consistent performance across EEG and random inputs raises concerns about whether the models are genuinely learning text-related information from EEG data. It suggests that they might be simply memorizing patterns from the training labels rather than extracting meaningful content from the EEG signals. To definitively determine this, it's crucial to compare model performance using EEG data against a baseline established with random noise input.
These findings underscore the need for more rigorous evaluation methodologies in EEG-to-Text research that avoid relying on teacher-forcing and ensure models are assessed based on their ability to learn from EEG data itself.

\section{Conclusion}

This study underscores the importance of meticulous evaluation practices in EEG-to-Text research. Our analysis revealed a critical shortcoming in previous research methodologies – the lack of a robust baseline for assessing model performance. By demonstrating that models perform similarly on pure noise compared to actual EEG data, we highlight the potential for inflated performance metrics when using teacher-forcing and previous evaluation methods. Our proposed methodology offers a more reliable way to assess a model's ability to extract meaningful information from EEG signals. It helps differentiate between models that truly learn from EEG data and those that might simply be memorizing patterns from training labels.

Looking ahead, we advocate for stricter evaluation practices within the EEG-to-Text research community. Researchers should ensure transparent reporting of methodologies and results, fostering open scientific discourse. Regularly incorporating noise baselines during evaluation is crucial for establishing a reliable performance benchmark. By employing more rigorous methodologies, researchers can shift the focus towards demonstrably learning text representations from EEG data, paving the way for advancements in this field. The adoption of these practices will lead to more reliable research findings and ultimately accelerate the development of robust and effective EEG-to-Text communication systems.

\begin{table}
\centering
\small 
\setlength{\tabcolsep}{3pt} 
\renewcommand{\arraystretch}{1.15} 
\caption{EEG-to-Text model evaluation on the ZuCo datasets, incorporating reading tasks from \textbf{SR v1.0, NR v1.0, and TSR v1.0}. "\textit{w/tf}" denotes results obtained using teacher-forcing during evaluation as utilized in the original study \cite{wang2022open_aaai_eeg2text}. In the training and evaluation phases, "EEG" denotes the use of word-level EEG features, while "Random" refers to the employment of random numbers generated from a normal distribution.}
\label{Results_TSRver}
\resizebox{\textwidth}{!}{
\begin{tabular}{ccccccccccc} 
\toprule
                        &                           &                                                  & \multicolumn{4}{c}{\textbf{BLEU-N (\%)}} & \multicolumn{3}{c}{\textbf{ROUGE-1 (\%)}} & \textbf{WER (\%)}     \\ 
\cline{4-11}
\textbf{\textbf{Pretrained model}} & \multicolumn{1}{c}{\textbf{\textbf{Training}}} & \multicolumn{1}{c}{\textbf{\textbf{Evaluation}}} & N=1   & N=2   & N=3   & N=4              & P     & R     & F                         & \multicolumn{1}{l}{}  \\ 
\midrule
BART                    & EEG                                         & EEG                                              & 13.69 & 2.97  & 0.82  & 0.32             & 11.98 & 13.43 & 11.87                     & 108.43                \\
                        & EEG                                         & Random                                           & 13.87 & 3.09  & 0.77  & 0.25             & 12.23 & 13.60 & 12.14                     & 108.31                \\
                        & Random                                      & EEG                                              & 14.05 & 3.12  & 1.00  & 0.41             & 11.46 & 12.37 & 11.14                     & 110.96                \\
                        & Random                                      & Random                                           & 14.22 & 3.06  & 0.93  & 0.39             & 11.62 & 12.29 & 11.19                     & 110.98                \\
\midrule
BART \textit{w/tf} \cite{wang2022open_aaai_eeg2text}              & EEG                                         & EEG                                              & 39.31 & 22.09 & 12.49 & 7.27             & 26.41 & 31.40 & 28.58                     & 78.08                 \\
                        & EEG                                         & Random                                           & 39.34 & 22.13 & 12.52 & 7.29             & 26.44 & 31.43 & 28.61                     & 78.07                 \\
                        & Random                                      & EEG                                              & 39.67 & 22.15 & 12.49 & 7.12             & 26.29 & 31.00 & 28.34                     & 78.09                 \\
                        & Random                                      & Random                                           & 39.69 & 22.17 & 12.50 & 7.12             & 26.32 & 31.03 & 28.37                     & 78.09                 \\ 
\midrule
\midrule
Pegasus                 & EEG                                         & EEG                                              & 8.47  & 2.48  & 0.81  & 0.25             & 0.00  & 0.00  & 0.00                      & 99.69                 \\
                        & EEG                                         & Random                                           & 8.58  & 2.48  & 0.78  & 0.00             & 0.00  & 0.00  & 0.00                      & 99.89                 \\
                        & Random                                      & EEG                                              & 9.12  & 2.70  & 0.91  & 0.23             & 0.00  & 0.00  & 0.00                      & 98.73                 \\
                        & Random                                      & Random                                           & 9.06  & 2.60  & 0.84  & 0.00             & 0.00  & 0.00  & 0.00                      & 99.24                 \\
\midrule
Pegasus \textit{w/tf}            & EEG                                         & EEG                                              & 38.18 & 21.04 & 11.50 & 6.09             & 26.72 & 30.51 & 28.38                     & 78.57                 \\
                        & EEG                                         & Random                                           & 38.30 & 21.09 & 11.57 & 6.12             & 26.84 & 30.65 & 28.51                     & 78.56                 \\
                        & Random                                      & EEG                                              & 39.10 & 21.74 & 11.97 & 6.17             & 27.43 & 31.26 & 29.11                     & 78.09                 \\
                        & Random                                      & Random                                           & 39.17 & 21.70 & 11.96 & 6.18             & 27.41 & 31.34 & 29.14                     & 78.10                 \\ 
\midrule
\midrule
T5                      & EEG                                         & EEG                                              & 16.64 & 5.80  & 1.96  & 0.81             & 12.28 & 12.88 & 11.85                     & 111.13                \\
                        & EEG                                         & Random                                           & 15.42 & 4.78  & 1.57  & 0.65             & 10.57 & 11.45 & 10.35                     & 112.00                \\
                        & Random                                      & EEG                                              & 15.95 & 5.71  & 2.01  & 0.91             & 11.90 & 12.61 & 11.47                     & 111.37                \\
                        & Random                                      & Random                                           & 15.54 & 5.22  & 1.70  & 0.67             & 11.48 & 12.23 & 11.10                     & 111.74                \\
\midrule
T5 \textit{w/tf}                 & EEG                                         & EEG                                              & 43.50 & 25.50 & 15.18 & 8.69             & 22.92 & 28.23 & 25.11                     & 81.39                 \\
                        & EEG                                         & Random                                           & 43.53 & 25.56 & 15.15 & 8.68             & 22.76 & 27.82 & 24.87                     & 81.43                 \\
                        & Random                                      & EEG                                              & 43.47 & 25.34 & 15.03 & 8.67             & 23.02 & 27.96 & 25.06                     & 81.64                 \\
                        & Random                                      & Random                                           & 43.63 & 25.57 & 15.23 & 8.78             & 23.36 & 28.40 & 25.45                     & 81.46                 \\
\bottomrule
\end{tabular}
}
\end{table}

\begin{table}
\centering
\small 
\setlength{\tabcolsep}{3pt} 
\renewcommand{\arraystretch}{1.15} 
\caption{EEG-to-Text model evaluation on the ZuCo datasets, incorporating reading tasks from \textbf{SR v1.0, NR v1.0, and NR v2.0}. "\textit{w/tf}" denotes results obtained using teacher-forcing during evaluation as utilized in the original study \cite{wang2022open_aaai_eeg2text}. In the training and evaluation phases, "EEG" denotes the use of word-level EEG features, while "Random" refers to the employment of random numbers generated from a normal distribution.}
\label{Results_NRver}
\resizebox{\textwidth}{!}{
\begin{tabular}{ccccccccccc} 
\toprule
                        &                           &                                                  & \multicolumn{4}{c}{\textbf{BLEU-N (\%)}} & \multicolumn{3}{c}{\textbf{ROUGE-1 (\%)}} & \textbf{WER (\%)}     \\ 
\cline{4-11}
\textbf{\textbf{Pretrained model}} & \multicolumn{1}{c}{\textbf{\textbf{Training}}} & \multicolumn{1}{c}{\textbf{\textbf{Evaluation}}} & N=1   & N=2   & N=3   & N=4              & P     & R     & F                         & \multicolumn{1}{l}{}  \\ 
\midrule
BART                    & EEG                       & EEG                                              & 11.58 & 3.40  & 1.33  & 0.54             & 11.33 & 15.44 & 12.40                     & 99.68                 \\
                        & EEG                       & Random                                           & 11.26 & 3.19  & 1.28  & 0.57             & 11.07 & 15.08 & 12.06                     & 100.03                \\
                        & Random                    & EEG                                              & 11.84 & 3.36  & 1.41  & 0.68             & 11.39 & 15.17 & 12.24                     & 100.88                \\
                        & Random                    & Random                                           & 11.78 & 3.18  & 1.23  & 0.52             & 11.22 & 15.05 & 12.07                     & 101.12                \\
\midrule
BART \textit{w/tf} \cite{wang2022open_aaai_eeg2text}               & EEG                       & EEG                                              & 41.22 & 24.18 & 13.87 & 7.77             & 29.52 & 36.31 & 32.42                     & 75.33                 \\
                        & EEG                       & Random                                           & 41.22 & 24.18 & 13.87 & 7.77             & 29.52 & 36.32 & 32.42                     & 75.31                 \\
                        & Random                    & EEG                                              & 41.63 & 24.70 & 14.44 & 8.45             & 29.56 & 35.83 & 32.30                     & 74.76                 \\
                        & Random                    & Random                                           & 41.61 & 24.69 & 14.40 & 8.40             & 29.55 & 35.77 & 32.27                     & 74.76                 \\ 
\midrule
\midrule
Pegasus                 & EEG                       & EEG                                              & 10.57 & 2.82  & 0.86  & 0.27             & 0.00  & 0.00  & 0.00                      & 99.21                 \\
                        & EEG                       & Random                                           & 11.14 & 2.75  & 0.88  & 0.30             & 0.00  & 0.00  & 0.00                      & 99.70                 \\
                        & Random                    & EEG                                              & 9.06  & 2.57  & 1.00  & 0.36             & 0.00  & 0.00  & 0.00                      & 98.37                 \\
                        & Random                    & Random                                           & 9.08  & 2.54  & 0.92  & 0.22             & 0.00  & 0.00  & 0.00                      & 98.40                 \\
\midrule
Pegasus \textit{w/tf}            & EEG                       & EEG                                              & 41.18 & 23.42 & 13.12 & 7.22             & 30.32 & 34.83 & 32.31                     & 75.30                 \\
                        & EEG                       & Random                                           & 41.06 & 23.17 & 12.82 & 6.86             & 30.08 & 34.59 & 32.07                     & 75.51                 \\
                        & Random                    & EEG                                              & 41.89 & 23.90 & 13.27 & 7.09             & 30.88 & 35.52 & 32.93                     & 74.67                 \\
                        & Random                    & Random                                           & 41.73 & 23.80 & 13.29 & 7.22             & 30.69 & 35.31 & 32.73                     & 74.88                 \\ 
\midrule
\midrule
T5                      & EEG                       & EEG                                              & 16.76 & 6.15  & 2.56  & 1.26             & 13.69 & 15.27 & 13.71                     & 104.50                \\
                        & EEG                       & Random                                           & 15.70 & 5.44  & 2.04  & 0.93             & 12.25 & 13.84 & 12.35                     & 106.01                \\
                        & Random                    & EEG                                              & 15.86 & 5.47  & 2.17  & 1.02             & 12.88 & 14.75 & 13.03                     & 104.30                \\
                        & Random                    & Random                                           & 16.26 & 5.76  & 2.23  & 1.03             & 13.58 & 15.44 & 13.66                     & 104.49                \\
\midrule
T5 \textit{w/tf}                 & EEG                       & EEG                                              & 46.03 & 28.23 & 17.35 & 10.55            & 28.14 & 33.84 & 30.53                     & 77.06                 \\
                        & EEG                       & Random                                           & 45.55 & 27.71 & 16.85 & 10.12            & 27.59 & 32.99 & 29.87                     & 77.63                 \\
                        & Random                    & EEG                                              & 45.62 & 27.78 & 16.79 & 10.15            & 27.26 & 32.93 & 29.65                     & 77.44                 \\
                        & Random                    & Random                                           & 45.75 & 27.96 & 16.99 & 10.21            & 27.39 & 33.05 & 29.77                     & 77.20                 \\
\bottomrule
\end{tabular}
}
\end{table}

\begin{table}[t]
\centering
\small 
\setlength{\tabcolsep}{3pt} 
\renewcommand{\arraystretch}{1.15} 
\caption{Decoding examples of EEG-to-Text models \cite{wang2022open_aaai_eeg2text}. "EEG" and "Random" represent that the model is trained and tested on word-level EEG features and random numbers, respectively. BART is used as the pretrained seq2seq model for all models. "\textit{w/tf}" denotes results obtained using teacher-forcing during evaluation, as utilized in the original study \cite{wang2022open_aaai_eeg2text}. \textbf{Bold} indicates words exactly matching the ground truth. \underline{Underline} denotes words consistently generated by models without teacher-forcing.}
\label{Text_generation}
\begin{tabular}{p{0.15\linewidth} p{0.82\linewidth}} 
\toprule
\textbf{Ground truth}       & It's not a particularly \textbf{good} film,\textbf{ but }neither \textbf{is it} a monsterous \textbf{one}.                                                 \\ 
\midrule
EEG \textit{w/tf} \cite{wang2022open_aaai_eeg2text}       & 's a a bad \textbf{good }movie, \textbf{but }it\textbf{ is it} bad bad. \textbf{one}.                                                                      \\
Random \textit{w/tf} & 's a a bad \textbf{good }movie, \textbf{but }it\textbf{ is it }bad bad. \textbf{one}.                                                                      \\
EEG                     & \underline{He was} elected to the United States House of Representatives in 1946.                                                                              \\
Random               & \underline{He was} educated at the University of Virginia, where he earned a Bachelor of Arts degree in political science.                                     \\ 
\midrule
\midrule
\textbf{Ground truth}       & Everything its title \textbf{implies}, a standard-\textbf{issue} crime \textbf{drama} spat out from the Tinseltown assembly \textbf{line}.                 \\ 
\midrule
EEG \textit{w/tf} \cite{wang2022open_aaai_eeg2text}      & about predecessor \textbf{implies} is and movie,\textbf{issue}, \textbf{drama}. between of the depthsseltown set \textbf{line}.                            \\
Random \textit{w/tf} & about predecessor \textbf{implies} is and movie,\textbf{issue}, \textbf{drama}. between of the sameseltown set..                                           \\
EEG                     & \underline{He was} a member of the Democratic National Committee (DNC) from 1952 until his death in 1968.                                                      \\
Random               & \underline{He was} educated at Trinity College, Cambridge and the University of Oxford.                                                                        \\ 
\midrule
\midrule
\textbf{Ground truth}       & Joseph H. Ball (November 3, 1905 - December 18, \textbf{1993) was an American} politician.                                                                 \\ 
\midrule
EEG \textit{w/tf} \cite{wang2022open_aaai_eeg2text}       & was. Smith (born 23, 18 - April 23, \textbf{1993) was an American }actor,                                                                                  \\
Random \textit{w/tf} & wasux W (born 23, 18 - April 23, 1977) \textbf{was an American} actor.                                                                                     \\
EEG                     & \underline{He was} elected to the United States House of Representatives in 1920.                                                                              \\
Random               & \underline{He was} educated at the University of Wisconsin-Madison, and received a Bachelor of Arts degree in English literature from Stony Brook University.  \\
\bottomrule
\end{tabular}
\end{table}

\clearpage

\medskip

{
\small
\bibliography{neurips_2024}
}


\end{document}